%% file: main.tex
\documentclass[conference,letterpaper,10pt]{IEEEtran}

\IEEEoverridecommandlockouts

\usepackage{cite}
\usepackage{amsmath,amssymb,amsfonts}
\usepackage{algorithmic}
\usepackage{graphicx}
\usepackage{textcomp}
\usepackage{xcolor}
\usepackage{url}
\usepackage{booktabs}
\usepackage{microtype}
\usepackage{siunitx}
\usepackage{balance}
\usepackage[hidelinks]{hyperref}

\graphicspath{{fig/}}

\newcommand{\code}[1]{\texttt{\small #1}}

\newcommand{\Real}{\mathbb{R}}
\newcommand{\Linf}{L_{\infty}}
\newcommand{\Ltwo}{L_{2}}

\title{Online Hamilton--Jacobi Reachability Onboard the Robot Control Loop:\\A Real-Time GPU Solver for\\Safety-Critical Autonomy}

\author{
  \IEEEauthorblockN{Kevin Weekly}
  \IEEEauthorblockA{
    Mirror AI \\
    \texttt{kevin.weekly@mirrorai.dev}
  }
  \and
  \IEEEauthorblockN{Alexandre M. Bayen}
  \IEEEauthorblockA{
    University of California, Berkeley \\
    \texttt{bayen@berkeley.edu}
  }
}

\begin{document}
\maketitle

\input{sections/abstract}


\input{sections/intro}
\input{sections/background}
\input{sections/related}
\input{sections/solver}
\input{sections/verification}
\input{sections/results}

\input{sections/applications}
\input{sections/discussion}
\input{sections/conclusion}
\input{sections/acknowledgment}

\balance
\bibliographystyle{IEEEtranDOI}
\bibliography{refs}

\end{document}

%% file: sections/abstract.tex
\begin{abstract}
Hamilton--Jacobi (HJ) reachability provides certified backward
reachable tubes (BRTs) for safety-critical robotic systems, but the
level-set computation scales exponentially in the state dimension and
has historically been performed offline. As autonomy stacks
(ADAS/AV, humanoid robotics, and vision--language--action models)
increasingly rely on black-box, probabilistic policies, a deterministic
safety layer with formal guarantees is an essential complement. We
present a CUDA implementation of the canonical WENO5 + TVD-RK3 +
Lax--Friedrichs level-set Hamilton--Jacobi--Isaacs solver for 2D, 3D,
and 4D dynamics that is fast enough to run inside a robot's control
loop. In 32-bit floating point (FP32) on a single workstation GPU
(NVIDIA RTX~4000 Ada), it computes the 3D Air3D BRT at $112\,$Hz, a 4D
pursuit--evasion BRT at $61\,$Hz, and a 2D wind-affected drifter BRT
at $74$--$147\,$Hz, $5.9$--$17.5\times$ faster than the JAX reference
\code{hj\_reachability} at matched precision and $1.7$--$16\times$
faster than a \code{torch.compile} reimplementation of the same
scheme. In 64-bit floating point (FP64) it agrees with
\code{hj\_reachability} on eight benchmark cases, reproduces Mitchell's
Air3D BRT to within $0.04\%$ of the reference cell count, and runs
$2.0$--$4.6\times$ faster. On an embedded NVIDIA Jetson AGX Thor it
computes Air3D at $66\,$Hz ($N{=}41$) and the 2D drifter at $54\,$Hz,
$3.8$--$8.9\times$ faster than \code{hj\_reachability} on the same
device. As a multi-agent application, an in-kernel label-propagation
extension recovers the $K$-agent capture-basin Voronoi tessellation
from a single combined HJ solve, independent of fleet size; it runs at
$4.0\,$ms per tick in a closed-loop wildfire-response simulation of six
quadcopters under time-varying, non-uniform wind.
\end{abstract}

%% file: sections/intro.tex
\section{Introduction}
\label{sec:intro}

The next generation of autonomous robotics --- advanced driver-assistance
and self-driving vehicles (ADAS/AV), humanoid platforms, and
vision--language--action (VLA) policies --- pushes decision-making
into large, learned, fundamentally probabilistically determined models whose worst-case
behavior is case-dependent and whose safety properties cannot be certified by
inspection. A deterministic, model-based safety layer with formal
guarantees is therefore not optional but a structural enabler of these
stacks: it is the mechanism by which a black-box upstream policy can
be trusted in safety-critical deployment, and the only available
complement that gives the overall system the worst-case behavior the
upstream model cannot. Hamilton--Jacobi (HJ) reachability is the
strongest available primitive for that role. The backward reachable
tube (BRT) is the exact set of states from which a constraint
violation cannot be avoided against the worst-case
disturbance~\cite{lygeros2004reach,mitchell2005tdhj,bansal2017survey}.
Three properties of the BRT make it the right object for safety, and
keeping them distinct avoids a common confusion. First, depending on
how it is computed, the BRT---or, more precisely, its boundary---delimits
the set of states from which, \emph{regardless of the control}, the
worst-case disturbance can force the system into the target set (here,
the unsafe set). Second, from any state outside the tube there always
exists a control that keeps the system safe over the horizon. Third, on
the boundary of the tube at least one control \emph{must} be applied to
stay safe: absent it, the disturbance can drive the system into the tube
and ultimately into the unsafe set.
A robot whose current state lies inside the BRT must either accept a
failure outcome or apply the value-function-derived optimal control
to escape, and so the BRT answers the most fundamental question a
safety-critical autonomy stack can ask~\cite{fisac2019safety,hsu2024safetyfilter}.
A BRT-based safety filter can sit downstream of a learned planner and
either certify its proposed action or override it with a provably
safe one~\cite{hsu2024safetyfilter,fisac2019safety}, giving the deployed
ADAS/AV, humanoid, or VLA stack a deterministic floor of behaviour
that the upstream learned policy is not by itself required to provide.

The well-known drawback is that the level-set computation that
delivers a guaranteed-fidelity BRT scales exponentially in the state
dimension. The historical response has been to compute the BRT once
offline and then use it as a fixed lookup in the control
loop~\cite{mitchell2008toolbox,helperoc,hjreachability}. This is a
critical limitation for next-generation autonomy: in any setting
where the disturbance model, the obstacle set, or the dynamics are
not known a priori --- the operating regime of every real ADAS
stack, every legged platform on unmodeled terrain, and every VLA
policy in an unfamiliar environment --- a stale offline BRT is a
stale safety guarantee.

With the implementation presented in this article, we demonstrate that the boundary between offline and online
BRT computation has moved. On a single workstation GPU (NVIDIA
RTX~4000 Ada), the canonical Mitchell~Air3D 3D pursuit--evasion
BRT~\cite{mitchell2005tdhj} solves in $9\,$ms at 32-bit floating point (FP32) with the same
numerical scheme used by the Stanford Autonomous Systems Laboratory (ASL) JAX reference
\code{hj\_reachability}~\cite{hjreachability}, a 4D pursuit--evasion
BRT at $31^4 \approx 10^6$ cells solves in $16\,$ms, and the 2D
drift-affected vehicle BRT solves at $74$--$147\,$Hz at WENO5 fidelity.
These rates place the level-set solver inside the inner control loop
of a robot, and shift the design question from how to live with a
single BRT precomputed for an entire mission to how to use a BRT that
is recomputed at every control tick.

Concretely, our solver implements the standard level-set HJI recipe on
CUDA---fifth-order WENO (WENO5) reconstruction of the spatial gradient,
total-variation-diminishing third-order Runge--Kutta (TVD-RK3) time
integration, and a Lax--Friedrichs numerical Hamiltonian---with one
fused kernel per RK3 stage (reconstruction, Hamiltonian, stage
combine, and BRT clamp in-register), a division-lean WENO5
reconstruction, a host-static CFL step, and an FP32 precision mode
that is the relevant regime for embedded robotic deployment. At
matched 64-bit floating point (FP64)
arithmetic (the academic-standard precision for verification studies)
our solver agrees with \code{hj\_reachability} on eight canonical
benchmarking cases at $2.0$--$4.6\times$ the speed on matched hardware,
reproduces Mitchell's Air3D BRT~\cite{mitchell2005tdhj} to within $0.04\%$ of the
reference cell count at $N{=}51$, and is $1.3$--$1.4\times$ faster
than a \code{torch.compile}~\cite{ansel2024pytorch2} reimplementation
of the identical scheme, a maximally-tuned tensor-compiler baseline.
In FP32 mode the solver is
$21$--$33\times$ faster than its own FP64 path with sub-grid BRT-set
deviation, and $5.9$--$17.5\times$ faster than the JAX reference at
matched FP32 on the same hardware. To verify that these rates are
not specific to a desktop-class GPU, we re-run the solver on an
NVIDIA Jetson AGX Thor embedded developer kit
and measure $27$--$66\,$Hz on the 3D Air3D and 2D drifter cases
($3.8$--$8.9\times$ faster than \code{hj\_reachability} on the same
device), with the closed-loop fleet solve drawing approximately $31\,$W
of total system input power. As a multi-agent
application we extend the 2D solver with an in-kernel
label-propagation kernel, in the spirit of competitive fast
marching~\cite{sethian1996fastmarching}, that recovers the $K$-agent capture-basin
Voronoi tessellation from a single combined HJ solve whose cost is
independent of fleet size, with
$99.5\%$ label agreement with a per-agent argmin reference.
We demonstrate the latter on a closed-loop wildfire-response fleet of
six quadcopters defending a $20 \times 20\,$km region under
time-varying, non-uniform wind.

The rest of this article is organized as follows.
Section~\ref{sec:background} reviews the HJ reachability formulation
and the level-set discretization recipe we use.
Section~\ref{sec:related} positions our solver against the
classical-fidelity offline tools and the learned-value online
approximators at the two ends of the existing literature.
Section~\ref{sec:solver} describes the CUDA implementation, the FP32
precision mode, and the boundary and obstacle postprocessors.
Section~\ref{sec:verification} describes the verification protocol
against \code{hj\_reachability}, and Section~\ref{sec:results}
reports cross-library agreement, the equi-accuracy comparison, and
the FP32 real-time envelope.
Section~\ref{sec:applications} presents the multi-agent
wildfire-response Voronoi demonstration.
Section~\ref{sec:discussion} discusses the dimensional ceiling,
the FP32 fidelity floor, embedded-platform performance on Jetson AGX
Thor, and future work. Section~\ref{sec:conclusion} concludes.

%% file: sections/background.tex
\section{Background}
\label{sec:background}

For a dynamical system $\dot x = f(x, u, d)$ on
$x \in \Real^d$ with control values $u \in \mathcal U$ and
disturbance values $d \in \mathcal D$, admissible control and
disturbance signals $u(\cdot) \in \mathbb{U}$ and $d(\cdot) \in \mathbb{D}$
(measurable maps from $[0, T]$ into $\mathcal U$ and $\mathcal D$,
respectively), and a failure set
$\mathcal{L} = \{x : \ell(x) \le 0\}$ described by a signed-distance
function $\ell$, the time-$T$ \emph{backward reachable tube} is
\begin{equation}
\label{eq:brt}
\begin{split}
\text{BRT}(T) = \{ x_0 : {} & \forall\, u(\cdot) \in \mathbb{U}\ \ \exists\, d(\cdot) \in \mathbb{D}\, \\
& \exists s \in [0, T]\ \ \ell(x_{x_0,u,d}(s)) \le 0 \},
\end{split}
\end{equation}
the set of initial states from which the disturbance can force entry
into $\mathcal L$ within the horizon regardless of the
control~\cite{lygeros2004reach,mitchell2005tdhj}. Mitchell, Bayen,
and Tomlin~\cite{mitchell2005tdhj} showed that the BRT is the
zero-sublevel set $\{x : V(t, x) \le 0\}$ of the value function $V$,
which satisfies the Hamilton--Jacobi--Isaacs (HJI) partial
differential equation
\begin{equation}
\label{eq:hji}
\partial_t V(t, x) + H(x, \nabla V(t, x)) = 0,
\qquad V(0, x) = \ell(x),
\end{equation}
integrated backward in time from the terminal condition
$V(0, x) = \ell(x)$~\cite{isaacs1965}, with the Isaacs-form Hamiltonian
\begin{equation}
\label{eq:ham}
H(x, p) = \max_{u \in \mathcal U}\, \min_{d \in \mathcal D}\, p^\top f(x, u, d)
\end{equation}
encoding the worst-case control--disturbance choice. The level-set
formulation then handles all of: minimum-time arrival to a target,
maximum-time avoidance of an obstacle, and
reach-while-avoid~\cite{mitchell2005tdhj,margellos2011reachavoid}.
Subsequent applications by Bayen, Mitchell, Oishi, and
Tomlin~\cite{bayen2007autolander} (aircraft autolander reach-set
analysis), by Weekly, Tinka, Anderson, and
Bayen~\cite{weekly2015hjriver} (HJ-based optimal-control navigation
of underactuated drifters under spatially-varying flow), and the
\emph{FaSTrack} framework of Herbert, Fisac, Tomlin, and
colleagues~\cite{herbert2017fastrack} (real-time guaranteed-safe
motion planning via a precomputed HJ tracking bound) demonstrated
that the same level-set primitive transfers to operational robotic
control once the fidelity-versus-runtime tradeoff is acceptable;
historically, that tradeoff has restricted the primitive to offline
mission planning.

We solve \eqref{eq:hji} on a uniform Cartesian grid by the standard
recipe~\cite{osher1991eno,jiang2000weno,shu1988tvd}: WENO5 spatial
reconstruction of $\nabla V$, TVD-RK3 time integration,
Lax--Friedrichs numerical Hamiltonian, and a CFL-limited time step.
We apply Mitchell's BRT-clamp $V \leftarrow \min(V_{\text{new}}, V_{\text{old}})$
in-kernel so that $V$ is monotonically non-increasing under
backward integration, the same role played by the
\code{backwards\_reachable\_tube} postprocessor of
\code{hj\_reachability}~\cite{hjreachability}.

%% file: sections/related.tex
\section{Related Work}
\label{sec:related}

\textbf{Classical level-set HJ tooling.}
Hamilton--Jacobi reachability was historically demonstrated at scale for control by
Mitchell's level-set
formulation~\cite{mitchell2002thesis,mitchell2005tdhj,osher1988level},
and the \emph{ToolboxLS} MATLAB
library~\cite{mitchell2008toolbox} together with its MATLAB
wrapper \emph{helperOC} and C++/CUDA port \emph{BEACLS}~\cite{helperoc} are still in
use today. \emph{OptimizedDP}~\cite{bui2022optimizeddp} compiles the
level-set HJ stencil to parallel multi-core CPU code via HeteroCL and
reports order-of-magnitude speedups over ToolboxLS, making higher state
dimensions tractable on the CPU. The Stanford ASL
\code{hj\_reachability} library~\cite{hjreachability} is a modern
JAX/XLA rewrite that exposes the same numerical primitives and is the
closest sibling of our solver; a PyTorch counterpart,
\code{hj\_reachability\_torch} of Choi et al.~\cite{choi2025},
provides similar primitives in eager PyTorch at first-order accuracy.
We treat the JAX library as the reference implementation in
Section~\ref{sec:results}. A like-for-like tensor-compiler comparison,
however, needs the \emph{identical} high-order scheme under
\code{torch.compile} --- which \code{hj\_reachability\_torch}, being
first-order and eager-mode, does not provide --- so we additionally
benchmark our own PyTorch reimplementation of the same
WENO5 + TVD-RK3 step compiled with
\code{torch.compile}~\cite{ansel2024pytorch2}, as a maximally-tuned
baseline.

\textbf{HJ for safe robotics.}
The reachability-based safety-filter pattern (i.e., run a nominal
controller and fall back to the HJ-optimal control when the safety
value function predicts imminent failure) was formalized
in~\cite{fisac2019safety,akametalu2014reach}, surveyed
in~\cite{bansal2017survey,chen2018reachsurvey,hsu2024safetyfilter},
and applied to aircraft autolander reach-set
analysis~\cite{bayen2007autolander}, HJ-based optimal-control
navigation of underactuated drifters under spatially-varying
flow~\cite{weekly2015hjriver}, fast tracking
(\emph{FaSTrack}~\cite{herbert2017fastrack}), robust
tracking under model mismatch~\cite{singh2020sos}, and
human-robot vehicle interactions~\cite{leung2020infusing}.
Bayen's line of work in
particular~\cite{mitchell2005tdhj,bayen2007autolander,weekly2015hjriver}
established the level-set HJ primitive as the right object for
operational reach-set computation in flow-affected and underactuated
robotics settings, and is the methodological lineage of the wind-affected
drifter cases and the wildfire-response demonstration we report in
Sections~\ref{sec:results} and~\ref{sec:applications}. All of these
works pay an offline computational cost for the BRT (typically
minutes to hours per case) and use it as a fixed lookup in the
control loop.

\textbf{Approximations to escape the dimension limit.}
DeepReach~\cite{bansal2021deepreach} and follow-up work train a
neural value function on the HJI residual to amortize the BRT
solve across many initial conditions. The trade-off is well known:
the neural value is fast to query but harder to certify, and the
training pass replaces the per-instance level-set
solve. Follow-up work conditions the learned value on parameters or
observations so the reachable set adapts
online~\cite{bansalParamCond,bansalObsCond} or refine an approximate
value function online via warm-started local grid HJ
updates~\cite{herbertOnline}, and related
efforts synthesize fast grid-based safety
filters~\cite{bahatiSafetyFilter}. Adversarial-actor-critic methods such as
ISAACS~\cite{hsu2023isaacs} take a complementary angle by learning a
robust controller against a learned worst-case disturbance.

\textbf{Grid-free and pointwise reachability.}
A complementary line of work sidesteps the grid entirely, solving
reachability pointwise through Hopf-type formulas rather than over a
dense array~\cite{darbonOsherHopf,chowDarbonOsher}. Recent methods extend
this to nonlinear systems by bounding the linearization error and treating
it as an adversary, recovering conservative guarantees at high state
dimension~\cite{sharplessLinEnvelopes,sharplessStateAug,sharplessKoopmanHopf,hopfreachability};
an iterative Hopf method evaluates the value pointwise without a
grid~\cite{leeIterHopf}, and ellipsoidal methods outer-approximate forward
reachable sets in real time~\cite{seoLeeFRS}. These approaches trade
the dense grid---and with it the full BRT surface---for fast pointwise
queries or conservative set approximations; our solver instead retains the full grid and makes the
whole-surface solve itself real-time.

\textbf{Predictive safety filters.}
A parallel literature builds MPC-based predictive safety filters
that guarantee constraint satisfaction without explicitly computing
the BRT~\cite{wabersich2021predictive,brunke2022safe}. These filters
trade a smaller online optimization for tighter modeling
assumptions; they are complementary to ours in that the BRT can
serve as the terminal safe set.

\textbf{Combined classical+sampling.}
\emph{DualGuard~MPPI}~\cite{borquez2025dualguard} pairs a sampling-based
MPC controller with a precomputed BRT used as a hard constraint, and is
the recent practitioner-oriented precedent closest in spirit to this
work. Our contribution is upstream of theirs: our solver delivers
the BRT fast enough for the precompute step to be replaced with an
online recompute, enabling the constraint to track shifts in the
disturbance model or robot dynamics during operation.
Sampling-based reachability estimators are another
practitioner-oriented approach in this vein~\cite{lew2021sampling}.

\textbf{Numerical foundations.}
Our discretization is standard:
WENO5~\cite{jiang2000weno,osher1991eno}, TVD-RK3~\cite{shu1988tvd},
and Lax--Friedrichs dissipation. The GPU implementation departs from
the classical shared-memory stencil
patterns~\cite{micikevicius2009gpu,datta2008stencil}: at control-rate
grid sizes the working set is L2-resident on current GPUs, and we
measure direct cache-served stencil fetches to be substantially
faster than shared-memory tiling (Section~\ref{sec:solver}).

\textbf{This work in context.}
The classical HJ tools provide provably convergent numerical approximation of the reachable set;
the approximation methods provide online queries with weaker
guarantees. Our solver occupies the underserved intersection:
classical-fidelity solves at robotics control time scales on commodity
GPU hardware. By \emph{robotics control time scales} we mean recomputing
the BRT within the time over which the robot's dynamics, constraints, or
disturbance model meaningfully change---on the order of tens of
milliseconds for an inner-loop safety filter, up to roughly a second for
a tactical dispatch loop---rather than once per mission.

%% file: sections/solver.tex
\section{Solver}
\label{sec:solver}

We integrate the Hamilton--Jacobi--Isaacs equation~\eqref{eq:hji}
backward from $t = 0$ to $t = -T$ over a uniform Cartesian grid in
$x \in \Real^d$. A possible terminal cost $\ell(x)$ is the signed distance to
the target or failure set, and the Isaacs-form Hamiltonian
$H(x, p)$~\eqref{eq:ham} encodes the worst-case dynamics under the
Isaacs formulation~\cite{mitchell2005tdhj,bansal2017survey}.
We compute the backward reachable tube (BRT), following Mitchell's approach, by clamping the
value after each time step, $V^{n+1} \leftarrow \min(V^{n+1}, V^{n})$, so
that $V$ is monotonically non-increasing in $|t|$. This plays the
role of \code{hj\_reachability}'s \code{backwards\_reachable\_tube}
Hamiltonian postprocessor, which clamps per RK substage instead
(Section~\ref{sec:verification}).

\subsection{Discretization}

Spatial derivatives in each direction are reconstructed with the
fifth-order weighted essentially non-oscillatory (WENO5) scheme of
Jiang and Peng~\cite{jiang2000weno}, using a six-cell stencil per
dimension. The numerical Hamiltonian uses Lax--Friedrichs
dissipation~\cite{osher1991eno}; we support global, local, and
local--local variants. Time integration uses the total-variation
diminishing third-order Runge--Kutta scheme of Shu and
Osher~\cite{shu1988tvd}, with the time step controlled to keep
$\Delta t \cdot \alpha_{\max} / \Delta x \le 0.75$.

\subsection{GPU kernel structure}

The 2D, 3D, and 4D solvers are implemented as CUDA kernels templated
on the floating-point type $\code{Real} \in \{\code{float}, \code{double}\}$
so that precision is selected at runtime via a single
\code{SolveOptions::precision} field. Each TVD-RK3 stage is a
\emph{single fused kernel}: per-axis WENO5 reconstruction, the
Lax--Friedrichs Hamiltonian, the RK3 stage combine, and (on the final
stage) the BRT clamp and value-postprocessors all execute in-register,
so the stage RHS never round-trips through global memory. The third
stage writes $V$ in place --- the stencil source is the second-stage
buffer, and the start-of-step value is only read pointwise --- which
also eliminates the per-step $V \!\to\! V_{\text{old}}$ copy. Stencil
values are fetched directly from global memory with boundary handling
(periodic wrap or linear extrapolation) inlined into the fetch: at the
grid sizes of interest the entire field is L2-resident on current
GPUs ($101^3$ FP32 is $4\,$MB against a $40\,$MB L2), which makes
both padded ghost-cell buffers and shared-memory tiles strictly
slower than letting the cache serve the stencil overlap: a
shared-memory-tiled 3D stencil pays a $5.4\times$ halo load
amplification, a block-wide barrier, and per-load boundary branching
for no benefit once the working set is L2-resident, and runs at
roughly one third the throughput of the direct-fetch kernel.

Two scheme-level invariants remove all per-step synchronization.
First, the WENO5 weight normalization is algebraically restructured
so that one reconstruction costs a single floating-point division
(the textbook form costs $13$): divides-by-constant become
compile-time reciprocal multiplies, and the three nonlinear weights
are cross-multiplied by the complementary smoothness denominators so
only the final normalization divides. This is identical up to
round-off and matters because consumer GPUs execute FP64 division in
software at a $1\!:\!64$ ALU ratio. Second, for every Hamiltonian in
this paper the Lax--Friedrichs wavespeed bound
$\alpha_{\max}$ is time-invariant and either constant or additively
separable in the state coordinates, so the CFL-limited $\Delta t$ is
computed once on the host instead of a full-grid wavespeed write plus
device reduction per RK stage. The 3D solver additionally captures
its whole step (three stage kernels plus a one-thread $\Delta t$
clamp) into a CUDA graph replayed once per step.

The first-order (RK1 + upwind) accuracy mode used by the real-time
fleet demonstration of Section~\ref{sec:applications} is fused the
same way: a single kernel per step does the direct-fetch upwind
derivatives, the Hamiltonian, the Euler update, the BRT clamp, the
postprocessors, and the time-to-reach update, ping-ponging between
two buffers with no per-step ghost-fill, reduction, or synchronization.
At the demonstration's $200^2$ grid the per-tick solve runs hundreds
of these steps, so reducing the step to a single launch is what keeps
the closed-loop solve inside its control budget (Section~\ref{sec:applications:result}).

\subsection{Boundary conditions and constraints}

Per-dimension boundary conditions are either Neumann
\emph{extrapolate} or \emph{periodic}; periodic dimensions use
the grid spacing $\Delta x = (x_{\mathrm{hi}} - x_{\mathrm{lo}})/N$, i.e., the
per-dimension spatial discretization of the domain
$[x_{\mathrm{lo}}, x_{\mathrm{hi}}]$ into $N$ cells, to wrap correctly. Two
value-postprocessors are exposed as drop-in options, matching the
semantics of \code{hj\_reachability}~\cite{hjreachability}:
\begin{itemize}
  \item \textbf{Static obstacle:}
        $V \leftarrow \max(V, g(x))$ where $g$ is the signed distance
        to the obstacle set.
  \item \textbf{Reach-while-avoid:}
        $V \leftarrow \min(\ell_t(x),\, \max(V, g(x)))$
        for target $\ell_t$ and obstacle $g$, the standard
        reach-while-avoid value function for differential
        games~\cite{margellos2011reachavoid}.
\end{itemize}
Both are applied by a single shared post-step kernel on the same
templated \code{Real}.

%% file: sections/verification.tex
\section{Verification Methodology}
\label{sec:verification}

We verify the solver against the Stanford ASL JAX library
\code{hj\_reachability}~\cite{hjreachability}, the de facto modern
reference for HJ reachability computation. Eight benchmarking cases
spanning 2D, 3D, and 4D dynamics are defined in shared
JSON specifications consumed by both libraries
(Section~\ref{sec:results}); each case is solved at three to
five grid resolutions per backend.

For each case we run both libraries with identical scheme
configuration: WENO5 spatial reconstruction, TVD-RK3 time
integration, \emph{global} Lax--Friedrichs dissipation (except on the
4D cases, see below), CFL factor
$0.75$, FP64 arithmetic, and the same signed-distance initial value
evaluated on the shared grid. Both libraries make $V$ monotonically
non-increasing under backward integration, but with different discrete
operators. Our \code{restrictUpdate} flag applies
$V^{n+1}_{\mathbf{m}} \leftarrow \min(V^{n+1}_{\mathbf{m}},\, V^{n}_{\mathbf{m}})$
once per full TVD-RK3 step, in the epilogue of the third stage, at
every grid node $\mathbf{m}$ (a multi-index over the $d$-dimensional
grid, with $n$ the time-step index). \code{hj\_reachability}'s
\code{backwards\_reachable\_tube} instead clamps the numerical
Hamiltonian, $\partial_t V \leftarrow -\min(\hat H, 0)$ in its sign
convention, inside each of the three Euler substages. The two
operators coincide for a forward-Euler step but not for TVD-RK3; the
agreement reported below is measured with this difference in place. Periodic dimensions are wrapped with the same
$\Delta x = (x_{hi} - x_{lo})/N$ convention on both sides. The
JAX reference is forced to FP64 via
\code{jax.config.update("jax\_enable\_x64", True)} so that the
single-call wall-clock measurement is comparable to ours-FP64; without
this, JAX silently defaults to FP32, whose ALU throughput on
non-datacenter Ada Lovelace silicon ($64\!:\!1$ over FP64 on the
RTX~4000 Ada) would understate JAX's FP64 time by an order of
magnitude.

Two scheme choices differ between the libraries on the 4D
pursuit--evasion cases, and together they account for the cross-library
gap there. First, our 4D kernel uses a gradient-local Lax--Friedrichs
coefficient, $\alpha_d = v\,|p_d|/\|p_{\text{pair}}\| \le v$, rather
than the global bound $\alpha_d = v$. Second, non-periodic dimensions
are extended by linear extrapolation in our solver, whereas
\code{hj\_reachability}'s default boundary condition extrapolates away
from zero; the 4D capture set touches the domain boundary, so this
choice is visible there (on Air3D the target is interior). With the
default settings our 4D BRT is $7.1$--$7.5\%$ smaller than the
reference at $N = 31$--$51$ at either precision. Switching the
reference to linear extrapolation shrinks its BRT by $5.6\%$, and
switching our kernel to the global coefficient grows ours by
$2.1$--$2.3\%$. With both aligned, the cell counts agree to within
$0.6\%$, $0.4\%$ and $0.3\%$ at $N = 31$, $41$ and $51$.

The reference postprocessor is set to
\code{hj.solver.backwards\_reachable\_tube}, which clamps the
Hamiltonian update so that $V$ is monotonically non-increasing under
backward integration. Without this clamp, \code{hj\_reachability}
computes the unclamped HJI value function, which permits $V$ to grow
at safe states and is a different object from the BRT. For
cases with static obstacles or reach-while-avoid semantics
(Section~\ref{sec:solver}), the reference value-postprocessor
composes the wall-mask, obstacle-clamp, and target-clamp stages in
the same order as ours.

For each $(\textsc{case}, N, \textsc{backend})$ tuple we record the
single-call wall-clock time, peak GPU memory via JAX
\code{memory\_stats()} or \code{cudaMemGetInfo}, and the full stack of
$V(t_k, \cdot)$ snapshots and times for downstream level-crossing
time-to-reach (TTR) extraction. The comparison driver computes
$\Linf$ and $\Ltwo$ differences pointwise over the interior domain,
TTR-field differences cell-by-cell, and convergence orders by
log-log regression of $\Linf$ disagreement versus $\Delta x$. All
hardware (NVIDIA RTX~4000 Ada, 20\,GB, driver~580.178.04) is shared
between backends to keep timing comparisons hardware-neutral.

%% file: sections/results.tex
\section{Results}
\label{sec:results}

The eight benchmarking cases (Table~\ref{tab:cases}) span 2D
holonomic and double-integrator dynamics, 2D drift-affected
autonomous vehicles, the canonical 3D Mitchell--Air3D
pursuit--evasion, and 4D two-vehicle pursuit--evasion with and
without state-space walls.

\begin{table}[t]
\centering
\caption{Benchmarking cases. Short labels used in Tables~\ref{tab:agreement}--\ref{tab:fp32}. Here $\|{\cdot}\|_2$ is the Euclidean norm and $|{\cdot}|$ the magnitude of a scalar control.}
\label{tab:cases}
\footnotesize
\setlength{\tabcolsep}{4pt}
\begin{tabular}{@{}lcl@{}}
\toprule
Case & State & Dynamics \\
\midrule
\code{holonomic\_l2}   & 2D & holonomic, $\|u\|_2 \le 1$ \\
\code{double\_int.}    & 2D & $f = [x_2, u]$, $|u| \le 1$ \\
\code{drifter\_const}  & 2D & constant-wind drifter \\
\code{drifter\_wg}     & 2D & spatial-wind drifter \\
\code{drifter\_obs}    & 2D & drifter + static obstacle \\
\code{air3d}           & 3D & Mitchell~Air3D~\cite{mitchell2002thesis} \\
\code{pe\_4d}          & 4D & 4D pursuit--evasion \\
\code{pe\_4d\_walls}   & 4D & 4D~PE~+~wall~mask \\
\bottomrule
\end{tabular}
\end{table}

\subsection{Cross-library agreement}
\label{sec:results:agreement}

We compare against two compiled GPU references: the JAX/XLA library
\code{hj\_reachability}, and our own PyTorch reimplementation of the
identical scheme (WENO5, TVD-RK3, Lax--Friedrichs, the same
ghost-cell and BRT-clamp operators) compiled with
\code{torch.compile}~\cite{ansel2024pytorch2}. Both are JIT-compiled, so the
comparison is compiled-against-compiled rather than an
eager-versus-compiled artifact, and both reproduce our BRT to the
tolerance summarized in Table~\ref{tab:agreement}: pointwise $\Linf$
disagreement $\le 0.04$ on the 2D cases and $\le 0.4$ in the
$V$-range $\approx 2.6$ for the 4D pursuit--evasion cases. This
disagreement is dominated by smearing of the non-smooth BRT boundary,
not smooth-region truncation error: all three apply the Mitchell
cumulative-min clamp, which introduces a non-differentiable front
that WENO5 cannot reconstruct at its formal fifth order --- a known
limitation of high-order WENO schemes for HJ
equations~\cite{mitchell2008toolbox}. Six of the eight cases are
distinct dynamics; the remaining two, \code{drifter\_obs} and
\code{pe\_4d\_walls}, reuse the \code{drifter\_const} and \code{pe\_4d}
dynamics and exist to verify the static-obstacle clamp and wall-mask
value-postprocessors that the reach-avoid and keep-out applications of
Section~\ref{sec:applications} depend on. They are checked for
accuracy here (Table~\ref{tab:agreement}) but, sharing the base-case
stencil, are excluded from the timing comparisons.

\begin{figure}[t]
\centering
\includegraphics[width=\columnwidth]{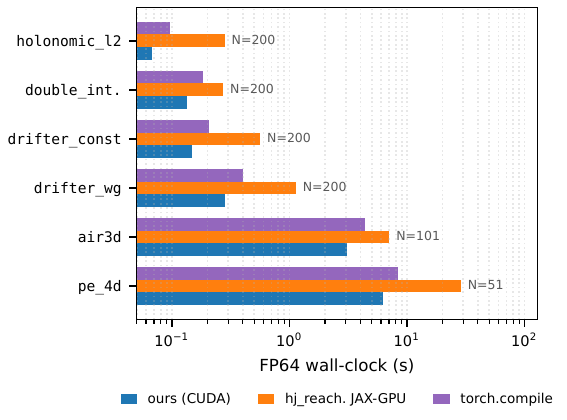}
\caption{FP64 single-call wall-clock at the largest matched grid for
the six distinct dynamics: our CUDA solver, the \code{hj\_reachability}
JAX-GPU reference, and our PyTorch reimplementation of the identical
WENO5+TVD-RK3 scheme compiled with \code{torch.compile}, on an NVIDIA
RTX~4000 Ada; medians of gated runs ($5$ ours, $3$ JAX, $5$
\code{torch.compile}). Ours is $2.0$--$4.6\times$ faster than JAX and
$1.3$--$1.4\times$ faster than \code{torch.compile}; the largest gains
over JAX appear on the drifter and
4D cases, where the fused per-stage kernels, in-kernel BRT clamping,
and host-static CFL step remove the most per-step work relative to
JAX/XLA's per-stage materialization. (The \code{drifter\_obs} and
\code{pe\_4d\_walls} postprocessor variants reuse the
\code{drifter\_const} / \code{pe\_4d} stencil, so they duplicate those
timings; their accuracy is in Table~\ref{tab:agreement}. JAX-CPU
wall-clock is in Table~\ref{tab:equi}.)}
\label{fig:speedup}
\end{figure}

Figure~\ref{fig:speedup} reports FP64 single-call wall-clock for the
six distinct dynamics at the largest matched grid. Our hand-written CUDA solver is
$2.0$--$4.6\times$ faster than JAX and $1.3$--$1.4\times$ faster than
\code{torch.compile}, both run at their best settings (ours:
\code{nvcc -O3 --use\_fast\_math}, \code{sm\_89}; \code{torch.compile}
at its default mode, which we found faster here than
\code{max-autotune}). The compiled baseline is
strong --- a tensor compiler emits many small, high-occupancy
elementwise kernels --- and we still lead it on every case by
exploiting scheme-level structure it cannot assume, analyzed in
Section~\ref{sec:discussion:torch}.

\begin{table}[!b]
\centering
\caption{Cross-library pointwise agreement at the largest matched grid.
\code{ours} versus \code{hj\_reachability} JAX-GPU, both at FP64,
WENO5 + TVD-RK3 + global Lax--Friedrichs dissipation, CFL $0.75$.
TTR ($\Linf$ in $\,$s) is derived from a level-crossing time
interpolation on the $V(t,x)$ snapshot stack.}
\label{tab:agreement}
\footnotesize
\setlength{\tabcolsep}{3pt}
\begin{tabular}{@{}lrSSS@{}}
\toprule
Case & $N$ & {$\Linf(V)$} & {$\Ltwo(V)$} & {$\Linf$(TTR, s)} \\
\midrule
\code{air3d}              & 101 & 0.0223  & 0.000396 & 0.0471 \\
\code{double\_int.}       & 200 & 0.00659 & 0.000109 & 0.0467 \\
\code{drifter\_const}     & 200 & 0.00439 & 0.000130 & 0.00548 \\
\code{drifter\_wg}        & 200 & 0.0130  & 0.000650 & 0.209  \\
\code{drifter\_obs}       & 200 & 0.0112  & 0.000442 & 0.0174 \\
\code{holonomic\_l2}      & 200 & 0.00217 & 0.000140 & 0.00157 \\
\code{pe\_4d}             & 51  & 0.347   & 0.0214   & 0.282  \\
\code{pe\_4d\_walls}      & 51  & 0.347   & 0.0219   & 0.550  \\
\bottomrule
\end{tabular}
\end{table}

\subsection{Air3D reproduction}
\label{sec:results:air3d}

\begin{figure}[t]
\centering
\includegraphics[width=\columnwidth]{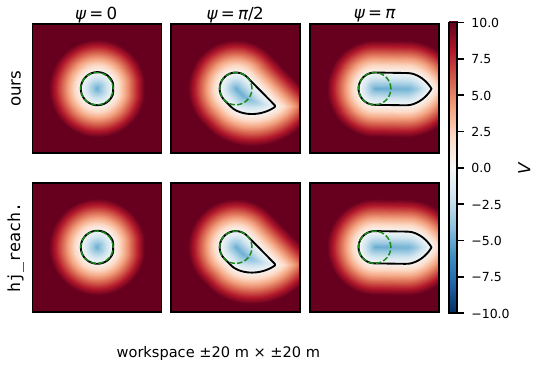}
\caption{Mitchell~Air3D BRT at $t=2\,\text{s}$ reproduced by our
solver (top row) and \code{hj\_reachability}~\cite{hjreachability}
(bottom row) at three relative-heading slices. Green dashed: the
$5\,\text{m}$ collision cylinder. Black contour: the BRT
boundary $\{V = 0\}$. BRT cell counts agree to within $0.04\%$ at
$N=51$.}
\label{fig:air3d}
\end{figure}

Figure~\ref{fig:air3d} reproduces Mitchell's canonical
3D pursuit--evasion BRT~\cite{mitchell2002thesis,mitchell2005tdhj}
at $t = 2\,\text{s}$ with $v_a = v_b = 5\,\text{m/s}$,
$\omega_{\max} = 1\,\text{rad/s}$, and a $5\,\text{m}$ collision
cylinder. Three relative-heading slices show the BRT boundary as the
zero level set of $V$. At $N = 51$ our solver and
\code{hj\_reachability} agree on the number of BRT cells to four cells
out of $\approx 11200$, or $0.04\%$.

\subsection{Equi-accuracy speedup}
\label{sec:results:equi}

Wall-clock at a fixed grid is one comparison; the other is to ask, at
the same target $\Linf$ accuracy, which library reaches it first.
Because both libraries converge to the same continuous solution as the
grid refines, their pointwise difference shrinks monotonically with
resolution. Table~\ref{tab:equi} reports, for each case, the
smallest grid at which the cross-library $\Linf$ falls below a
target $\varepsilon$, and the wall-clock at that grid. The relative
ranking is unchanged: ours is $3.1$--$4.9\times$ faster at the same
target accuracy on every case.

\begin{table}[t]
\centering
\caption{Equi-accuracy speedup. Wall-clock at the smallest grid where
$\Linf(V_{\text{ours}} - V_{\text{ref}}) \le \varepsilon$ for each
backend. All FP64; medians of gated runs ($5$ ours, $3$ JAX-GPU, $3$
JAX-CPU).}
\label{tab:equi}
\scriptsize
\setlength{\tabcolsep}{3pt}
\begin{tabular}{@{}lcrrrr@{}}
\toprule
Case & $\varepsilon$ & ours-GPU & JAX-GPU & JAX-CPU & vs \\
     &               & (s)       & (s)      & (s)      & JAX-GPU \\
\midrule
\code{air3d}            & 0.05 & 0.0969 & 0.362  & 1.15   & $3.7\times$ \\
\code{double\_int.}     & 0.02 & 0.00929 & 0.0285 & 0.0386 & $3.1\times$ \\
\code{drifter\_const}   & 0.02 & 0.0102 & 0.0428 & 0.0354 & $4.2\times$ \\
\code{drifter\_wg}      & 0.05 & 0.0205 & 0.0766 & 0.0978 & $3.7\times$ \\
\code{holonomic\_l2}    & 0.01 & 0.00475 & 0.0233 & 0.0283 & $4.9\times$ \\
\code{pe\_4d}           & 0.4  & 0.543  & 2.02   & 10.8   & $3.7\times$ \\
\code{pe\_4d\_walls}    & 0.4  & 0.550  & 2.03   & 10.8   & $3.7\times$ \\
\bottomrule
\end{tabular}
\end{table}
These FP64 wall-clocks establish cross-library parity for verification;
the rates relevant to closed-loop robot operation are the single-call
FP32 times of Section~\ref{sec:results:fp32} and
Figure~\ref{fig:envelope}.

\subsection{Real-time FP32 mode}
\label{sec:results:fp32}

On non-datacenter Ada Lovelace silicon FP64 runs at a small
fraction of the FP32 rate; the peak ALU ratio on the RTX~4000
Ada is $64\!:\!1$. With the division-lean reconstruction of
Section~\ref{sec:solver} the FP64 path is limited by FP64 arithmetic
throughput while the FP32 path is cache- and bandwidth-limited, and
the realized FP32 advantage is $21$--$33\times$, below the
peak ratio because the FP32 path is no longer arithmetic-bound. FP32 is in any case the
practical mode for closed-loop robotics where a BRT must refresh at
$10$--$100\,$Hz. Table~\ref{tab:fp32} reports the FP32 single-call
wall-clock of our solver against both compiled references on
identical hardware --- \code{hj\_reachability} (JAX/XLA) and our
\code{torch.compile} reimplementation --- together with the
achievable BRT-refresh rate. At FP32 our solver is
$5.9$--$17.5\times$ faster than JAX and $1.7$--$16\times$ faster than
\code{torch.compile}, the widest \code{torch.compile} margins falling
on the small, cache-resident 2D grids that fit a real-time budget.
At FP32 our BRT is set-identical to our FP64 BRT on the 2D and 3D
cases and within $0.1\%$ of its cells on the 4D cases
(Section~\ref{sec:discussion:fp32}); the cross-library 4D gap is a
scheme difference, not a precision effect (Section~\ref{sec:verification}).

\begin{table}[t]
\centering
\caption{FP32 single-call wall-clock and BRT-refresh rate on the
NVIDIA RTX~4000 Ada, against both compiled references:
\code{hj\_reachability} (JAX/XLA) and our \code{torch.compile}
reimplementation of the identical scheme. \emph{rate} is
$1/t_{\text{ours}}$ rounded to the nearest integer Hz. \emph{ours} and
JAX are medians of $9$ and $5$ single-call runs (driver 580.178.04),
each run gated on an idle host and GPU; the 3D solver replays its
CUDA-graph step without per-step host synchronization
(Section~\ref{sec:solver}). \code{torch.compile} (PyTorch 2.6.0,
TorchInductor default mode) is the median of $5$ timed solves after
compilation, gated the same way. All recover
the same BRT to the tolerance of Table~\ref{tab:agreement}.}
\label{tab:fp32}
\footnotesize
\setlength{\tabcolsep}{4pt}
\begin{tabular}{@{}lrrrrr@{}}
\toprule
Case & $N$ & ours & rate & JAX & torch.comp. \\
     &     & (ms) & (Hz) & (ms) & (ms) \\
\midrule
\code{air3d}          & 51  & 8.91   & 112  & 52.3   & 50.0 \\
\code{air3d}          & 101 & 102    & 9.8  & 804    & 178  \\
\code{drifter\_const} & 200 & 6.82   & 147  & 86.4   & 107  \\
\code{drifter\_wg}    & 200 & 13.6   & 74   & 159    & 223  \\
\code{pe\_4d}         & 31  & 16.4   & 61   & 150    & 27.6 \\
\code{pe\_4d}         & 51  & 222    & 4.5  & 3885   & 573  \\
\bottomrule
\end{tabular}
\end{table}

\begin{figure}[t]
\centering
\includegraphics[width=\columnwidth]{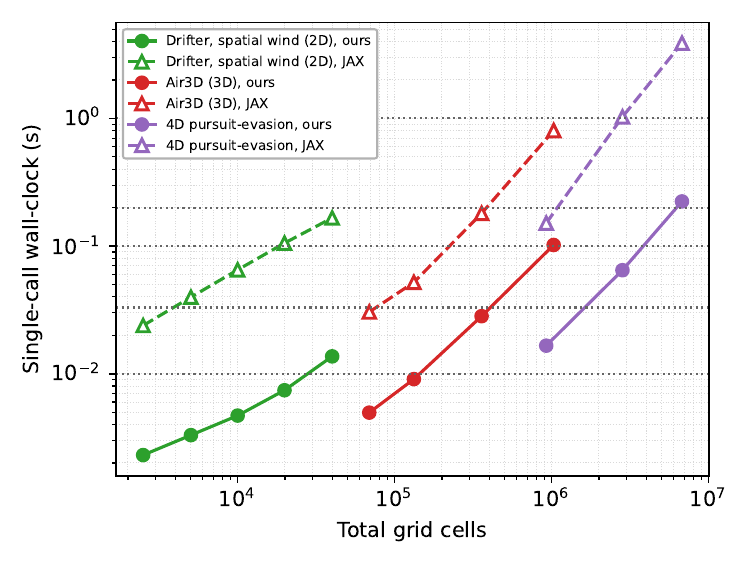}
\caption{Real-time operating envelope (the largest grid solvable within each control-rate budget): single-call FP32 wall-clock versus total
grid cell count for three representative cases. Solid lines with filled
markers are our solver; dashed lines with open markers are
\code{hj\_reachability} on the same hardware. Horizontal bands mark
the $100\,$Hz ($10\,$ms), $30\,$Hz ($33\,$ms), $10\,$Hz ($100\,$ms),
$5\,$Hz ($200\,$ms), and $1\,$Hz ($1\,$s) closed-loop control
budgets. The crossing of each curve with a budget line is the
largest grid at which the BRT can be recomputed every control tick
on commodity GPU hardware.}
\label{fig:envelope}
\end{figure}

Figure~\ref{fig:envelope} casts the same data as a real-time
\emph{envelope}: single-call FP32 wall-clock as a function of total
grid cells, for both our solver and the JAX reference, with
horizontal bands at the $100\,$Hz ($10\,$ms), $30\,$Hz ($33\,$ms),
$10\,$Hz ($100\,$ms), $5\,$Hz ($200\,$ms), and $1\,$Hz ($1\,$s)
closed-loop control budgets. Every point is a gated median ($7$ runs
ours, $3$ JAX) over the resolution sweep of
Section~\ref{sec:verification}. Our FP32 solver
is inside the $30\,$Hz
budget for Air3D up to $N = 71$ ($3.6\!\times\!10^5$ cells,
$36\,$Hz), 4D pursuit--evasion up to $N=31$ ($9.2\!\times\!10^5$
cells, $61\,$Hz), and the 2D drifter up to $N = 200$
($4\!\times\!10^4$ cells, $74$--$147\,$Hz, WENO5+RK3); the $100\,$Hz
budget covers Air3D to $N = 51$ ($8.9\,$ms) and the constant-wind
drifter to $N = 200$. Against \code{hj\_reachability} at matched
FP32, our solver is $5.7$--$17\times$ faster at every grid of the
sweep, with the
largest ratios on the wind-affected drifter and 4D pursuit--evasion
cases. JAX/XLA does not aggressively schedule FP32 stencil ops for
this style of compute, so the FP32-mode ratio is wider than the
$3.1$--$4.9\times$ FP64-mode ratio of Table~\ref{tab:equi}. The
crossing of each case curve with each control-rate line is the grid
at which an online safety filter at that rate becomes feasible on
commodity GPU hardware.

%% file: sections/applications.tex
\section{Online HJ Capture-Basin Voronoi for a Quadcopter Wildfire-Response Fleet}
\label{sec:applications}

\begin{figure*}[!t]
\centering
\includegraphics[width=\textwidth]{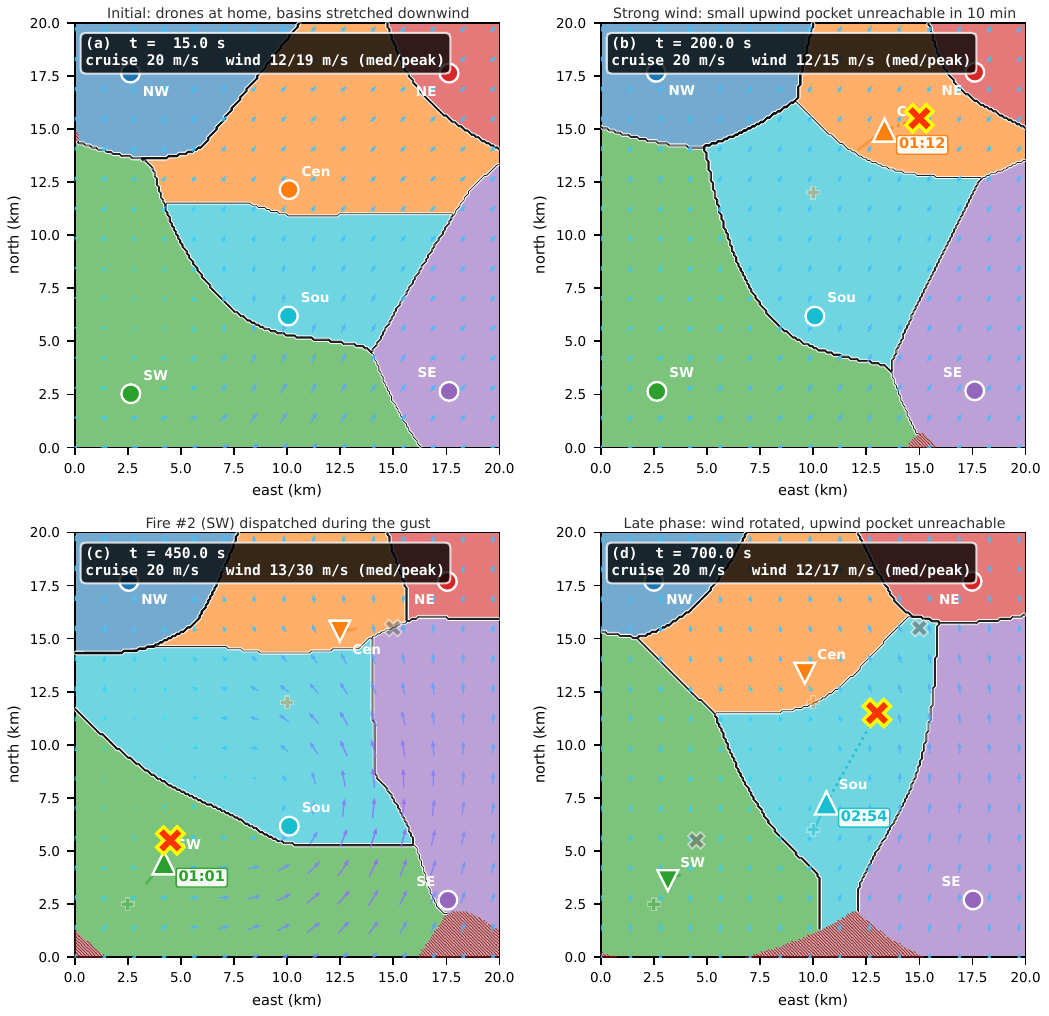}
\caption{Closed-loop wildfire-response fleet under time-varying,
non-uniform wind. Six homogeneous quadcopters defend a $20 \times 20\,$km
region (cruise $a = 20\,$m/s) against three fires. The joint HJ
capture-basin Voronoi tessellation, computed via in-kernel label
propagation from each drone's current position, is refreshed at every
$10\,$Hz outer-loop tick. Colored basins are each drone's $10$-minute
reachable region; red diagonal hatch marks cells unreachable within
the horizon by any drone. White curves are basin bisectors. Triangles
mark drones en route (with the $MM\!:\!SS$ predicted-arrival
countdown); inverted triangles, returning; circles, patrolling at
home. Red X marks a burning fire, gray X a suppressed fire. (a)
Initial state: drones at homes, basins anisotropic with the prevailing
NE wind. (b) The Center drone en route to Fire \#1 (NE
region); its basin has followed it east through the domain. (c) Fire
\#2 (SW region) en route during the gust, with the wind rotated.
(d) Late phase ($t = 700\,$s), wind rotated further: a pocket in the
upwind direction is unreachable in the $10$-min window.}
\label{fig:wildfire}
\end{figure*}

The compute-budget numbers in Section~\ref{sec:results:fp32} unlock a
problem class that has stayed offline: multi-agent operational
dispatch under time-varying, non-uniform disturbance. The
multi-agent control literature has long used Voronoi-style partitions
as the basic dispatch primitive --- coverage control via Voronoi
partitions of agent positions~\cite{bullo2005coverage}, Apollonius and reach-avoid game
geometry for pursuit--evasion~\cite{isaacs1965,garcia2019apollonius,yan2018reachavoid}
--- but these constructions are either closed-form only under
uniform disturbance or computed once per mission from fixed agent
homes. We instantiate the time-varying, spatially-varying-flow
extension here, namely a homogeneous fleet of quadcopters tasked
with suppressing wildfires in a $20 \times 20\,$km region under
realistic shifting wind, and show that the joint HJ time-to-reach
Voronoi tessellation can be recomputed at robot rates from the
fleet's \emph{live} positions, so that the dispatch decision reflects
the actual current state of the fleet (including drones already in
transit).

\subsection{Problem statement}
\label{sec:applications:problem}

Let $\{p_i(t)\}_{i=1}^{K}$ be the positions of $K$ homogeneous
holonomic agents (i.e., quadcopters in plan view) with dynamics
$\dot p = a u + w(t, p)$, $\|u\|_2 \le 1$, cruise speed $a$, and known
time-varying spatial wind $w(t, x)$. Given a \emph{tactical dispatch
horizon} $T$, the operational question at every control tick is:
which drone can reach each cell first, and which cells are
unreachable by any drone within $T$? We define the forward
time-to-reach (TTR) field for agent $i$ via the level-set problem
\begin{equation}
V_{i,t} + w(t,x) \!\cdot\! \nabla V_i + a \|\nabla V_i\| = 0,
\quad V_i(0, x) = \|x - p_i\| - r_d,
\label{eq:drifterTTR}
\end{equation}
where $r_d$ is the agent footprint. The capture-basin Voronoi
tessellation under horizon $T$ is then
$\Omega_i = \{x : V_i(T, x) \le 0,\ V_i(T, x) \le V_j(T, x)\,\forall j \neq i\}$
intersected with the reachable set
$\mathcal{R}(T) = \{x : \min_i V_i(T, x) \le 0\}$,
and the unreachable complement is
$\mathcal{U}(T) = \{x : V_i(T, x) > 0\,\forall i\}$.

\subsection{Why a level-set treatment is required}
\label{sec:applications:why}

For \emph{uniform} wind on an unbounded domain, the bisector between
two agents reduces to a classical anisotropic Apollonius-like curve.
Apollonius circles and their generalizations are the canonical
geometric primitive in pursuit--evasion differential
games~\cite{isaacs1965,garcia2019apollonius,yan2018reachavoid}, and
their equal-speed special case, the Euclidean Voronoi partition,
underlies multi-agent coverage control~\cite{bullo2005coverage}. Spatially-varying wind
eliminates that closed form: the time-optimal trajectory between two
points is no longer straight, the bisector curves in ways that follow
the curl and divergence of $w(t,x)$, and the unreachable-set boundary
depends on the global wind field through the Eikonal
equation~\cite{sethian1996fastmarching,tsitsiklis1995shortpaths}.
Numerical level-set propagation under the actual $w$ is the only
general method, and is the same primitive Weekly et
al.~\cite{weekly2015hjriver} used for underactuated drifter
navigation in spatially-varying flow.

\subsection{In-kernel label propagation: $K$ agents at the cost of one solve}
\label{sec:applications:labels}

The naive approach, i.e., $K$ independent TTR solves followed by an
argmin postprocess, scales linearly in fleet size. For homogeneous
agents we exploit the min-preservation property of HJ semigroups:
the combined field
$V(t, x) := \min_i V_i(t, x)$
satisfies the same HJ equation~\eqref{eq:drifterTTR} as each
individual $V_i$, with initial condition
$V(0, x) = \min_i (\|x - p_i\| - r_d)$. A single combined solve
produces $V$. To recover the assignment
$L(x) := \arg\min_i V_i(T, x)$, we propagate an integer label field $L$
\emph{inside the solver kernel}: at every time step, each cell whose
front has just arrived (i.e., $V$ has just crossed zero) inherits the
label of its 4-neighbor with the smallest $V$ among labeled
neighbors. That neighbor is the cell from which the front delivered
the assignment. The labeled combined solve costs roughly
$1.3\times$ a single-agent solve. Cells where $V > 0$ at the end of the solver
horizon stay at a sentinel value and constitute $\mathcal{U}(T)$. A
$3 \times 3$ majority postprocessor on the label field at save time
removes residual 1-cell discretization artifacts along bisectors.

This reduces $K$-agent capture-basin Voronoi to the cost of a single
TTR solve regardless of fleet size, provided the agents share
dynamics. We validated the combined-solve labels against ground-truth
argmin over $K$ independent solves at $K = 6$ on a $200 \times 200$
grid: the combined solve takes $15$--$17\,$ms versus $69$--$77\,$ms for
the six separate solves on the RTX~4000 Ada (calm and gusty wind,
medians of $7$ gated runs), a $4.6\times$ speedup against
the theoretical $K\times$ ceiling. Label agreement with the argmin
reference is $99.5\%$; the $0.5\%$ disagreement is a
$1$-cell stripe along Voronoi bisectors. To our knowledge this is the
first reported in-kernel Voronoi-from-HJ implementation in a
level-set HJ solver at GPU rates (the label kernel supports both the
WENO5+TVD-RK3 and the first-order modes; the results here use the
first-order mode of Section~\ref{sec:solver}). Competitive
fast-marching for Voronoi tessellation is
classical~\cite{sethian1996fastmarching}, but to date has not been
done inside an inner-loop-rate level-set HJ solver.

\subsection{Demonstration setup}
\label{sec:applications:setup}

We simulate $K = 6$ quadcopters at four perimeter and two interior
staging stations on the $20 \times 20\,$km region. Cruise speed
$a = 20\,$m/s. The wind field $w(t, x)$ combines (i) a synoptic
background rotating from NE to NW over the $15$-minute simulation at
$12\,$m/s magnitude (within the documented active-fire-spread range
of $8.3$--$27.7\,$m/s) and (ii) a cyclonic plume rotor whose center
walks slowly NE and whose tangential speed peaks at $30\,$m/s during
an $80$-second gust window centered at $t = 500\,$s, matching the
documented Santa-Ana extreme-gust regime. The tactical horizon is
$T = 10\,$min. Three fires appear at scripted times $t \in
\{60, 360, 660\}\,$s at scattered locations.

The dispatch policy proceeds as follows. When a fire appears we look
up the label field $L$ at the fire's cell. If the cell is unreachable
within the horizon $T$ (i.e., the label is the invalid sentinel), no
drone is dispatched and the fire stays pending until either the wind
shifts or a drone repositions such that some basin covers the cell.
If the live owner of the cell is currently suppressing another fire,
the new fire likewise waits. We do not include a fallback to
nearest-by-straight-line-distance dispatch. Note that a drone whose
10-minute basin does not currently cover the fire is, by definition,
not going to arrive at the fire within 10~minutes under the assumed
wind field, so such a fallback would assign a drone we already know
cannot complete the task within $T$.

\subsection{Closed-loop result}
\label{sec:applications:result}

The combined solve takes $4.0\,$ms per tick (mean over the
$9000$-tick run; median of $5$ gated runs) on the RTX~4000 Ada at
FP32 under the $10$-minute
horizon, giving an approximately $250\,$Hz Voronoi-refresh ceiling
against the $10\,$Hz outer loop. All three fires are suppressed
within the simulation window.
Figure~\ref{fig:wildfire} shows four representative frames.

The qualitative payoff is visible in
Figure~\ref{fig:wildfire}(b--d). As the Center drone leaves home to
suppress the NE fire, its basin (orange) extends east through the
domain following the drone, so any fire appearing in the central-east
region during that transit would be assigned to the Center drone
rather than to whichever drone's home station happened to be closest.
In practice, this is what it means for the Voronoi partition to be
recomputed from live positions rather than from static homes: the
partition tracks the fleet as it moves.

\subsection{Comparison with classical multi-agent dispatch}
\label{sec:applications:classical}

Voronoi partitioning of agent positions is the basic primitive of
coverage control~\cite{bullo2005coverage}; its common dispatch
instantiation is a static-home Voronoi (Euclidean or wind-anisotropic,
computed once per mission from each drone's home station). It
is closed-form, but cannot accommodate in-transit drones or
time-varying wind without re-solving, which puts it back in the same
regime as ours. Distributed sampling-based path planners (e.g.,
RRT$^\ast$ and its relatives) handle anisotropic flow but do not
produce a globally consistent partition. Linear MPC over the fleet
handles bounded disturbance, but the curse of dimensionality
precludes joint multi-agent optimization in a time-varying flow
field. Online HJ via in-kernel Voronoi delivers the global partition
under arbitrary $w(t, x)$ at robot rates, and the unreachable-set
$\mathcal{U}(T)$ emerges from the same solve at no extra cost.

%% file: sections/discussion.tex
\section{Discussion and Future Work}
\label{sec:discussion}

\subsection{The 4D ceiling}
\label{sec:discussion:4d}

The exponential scaling in state dimension is unchanged by this
work. At FP32 a $51^4 \approx 6.8 \cdot 10^6$-cell 4D solve takes
$0.22\,$s, nearly $7\times$ the $30\,$Hz control
budget; the $31^4 \approx 9.2 \cdot 10^5$-cell solve fits in
$16\,$ms but with coarse spatial resolution. State-space
decomposition and learned-value
approximations~\cite{herbert2017fastrack,bansal2021deepreach}
remain the appropriate tools above 4D; we only move the constants.

\subsection{FP32 fidelity floor}
\label{sec:discussion:fp32}

On the 2D and 3D cases the FP32 BRT is set-identical to ours-FP64
to the last cell; on the 4D pursuit--evasion cases it differs by up
to $0.1\%$ of cells, well below the inherent cross-library
disagreement at the same resolution ($\sim 0.35$ in a $V$-range of
$\sim 2.6$). A thin band along the BRT boundary will flip
safe/unsafe relative to the FP64 ground truth and should be treated
as a band of uncertain classification rather than a hard boundary.

\subsection{Embedded platform performance}
\label{sec:discussion:embedded}

The desktop benchmarks use an RTX~4000 Ada Generation GPU
($20\,$GB GDDR6 ECC, $130\,$W TDP, \code{sm\_89}). For deployment we
target an NVIDIA Jetson AGX Thor developer kit (T5000 module:
2560-core Blackwell GPU, $128\,$GB LPDDR5X unified memory,
\code{sm\_110}) at the $120\,$W power-mode cap. The same source
builds for both architectures with a single code path: the fused
per-stage kernels of Section~\ref{sec:solver} carry no
architecture-conditional dispatch, the CFL-limited time step is
either host-static or kept on device by a single-thread
\code{compute\_dt} kernel, and the 3D RK3 step is captured into a
CUDA graph and replayed back to back, with no per-step host
synchronization, on both devices.

On single solves Thor is $2.7$--$3.5\times$ slower than the desktop
(Table~\ref{tab:thor}): Air3D at $66\,$Hz at $N{=}41$ and $35\,$Hz at
$N{=}51$, the 2D drifter at $N{=}200$ at $54\,$Hz (constant wind) and
$27\,$Hz (spatially-varying wind), and 4D pursuit--evasion at
$N{=}31$ at $19\,$Hz, real-time for a typical
$5$--$20\,$Hz robot outer loop. The
multi-agent wildfire Voronoi
solve (Section~\ref{sec:applications}, $N{=}200$, in-kernel labels)
runs at $9.69\,$ms per tick on Thor ($103\,$Hz, mean over the
$9000$-tick closed-loop simulation, $3$ runs) with clocks pinned, all
three fires suppressed in every run; total system input power on the
Jetson VIN rail averaged $34.4\,$W (sampled at $1\,$Hz by
\code{tegrastats}). Under the default dynamic clocking the same run
(three runs, not idle-gated, host load average $0.3$--$1.0$) takes
$10.35\,$ms per tick at $31.1\,$W, the figure we quote for
deployment. Against the same \code{hj\_reachability}
reference on the same device, our solver runs $3.8$--$8.9\times$
faster at FP32. XLA's autotuner cannot run on \code{sm\_110} with
jaxlib~0.10.0 (its profiling kernels have no \code{sm\_110} image), so
the Thor JAX column uses XLA's default, untuned fusion configurations
(\code{--xla\_gpu\_autotune\_level=0}).

\begin{table}[h]
\centering
\caption{Measured wall-clock on Jetson AGX Thor ($120\,$W power-mode
cap, FP32), versus \code{hj\_reachability} JAX-GPU on the same
device, with the desktop-class RTX~4000 Ada ($130\,$W TDP, FP32)
baseline for context. Thor: $7$-run medians of \code{single\_call\_s}
with GPU clocks pinned (GPC $1386\,$MHz), each run gated on an idle
host; JAX is untuned XLA (see text). Ada: the gated medians of
Table~\ref{tab:fp32} ($9$ runs).
\textsuperscript{\dag}\code{fleet\_voronoi} is the closed-loop
Voronoi+TTR per-tick mean of Section~\ref{sec:applications}
($9000$-tick run; median of the per-run means over $3$ runs on Thor
and $5$ gated runs on Ada).}
\label{tab:thor}
\footnotesize
\setlength{\tabcolsep}{3pt}
\begin{tabular}{@{}lrrrrr@{}}
\toprule
Case & $N$ & Ada-ours & Thor-ours & Thor-JAX & ours/JAX \\
     &     & (ms)     & (ms)      & (ms)     & on Thor   \\
\midrule
\code{air3d}             & 41  & 4.95 & 15.1 & 68.5 & $4.5\times$ \\
\code{air3d}             & 51  & 8.91 & 28.9 & 110  & $3.8\times$ \\
\code{air3d}             & 101 & 102  & 355  & 1848 & $5.2\times$ \\
\code{drifter\_const}    & 200 & 6.82 & 18.6 & 167  & $8.9\times$ \\
\code{drifter\_wg}       & 200 & 13.6 & 37.1 & 289  & $7.8\times$ \\
\code{pe\_4d}            & 31  & 16.4 & 53.9 & 392  & $7.3\times$ \\
\code{fleet\_voronoi}\textsuperscript{\dag} & 200 & 4.04 & 9.69 & -- & -- \\
\bottomrule
\end{tabular}
\end{table}

\subsection{Hand-written kernels versus a tensor compiler}
\label{sec:discussion:torch}

Our lead over \code{torch.compile} (Table~\ref{tab:fp32},
Fig.~\ref{fig:speedup}) comes from \emph{scheme-level} information
that a general-purpose tensor compiler cannot assume. The compiler's schedule --- many small,
low-register, high-occupancy elementwise kernels that materialize
intermediates in memory --- is competitive with a kernel that is
merely fused, but four properties of the discretization let a
hand-written kernel go further. First, the WENO5 weight normalization
can be cross-multiplied into a single division because the weights
appear only as a convex ratio; the change is exact up to round-off,
but it is an algebraic identity, not a graph rewrite a compiler may
apply under IEEE semantics, and it matters because consumer Ada runs
FP64 at a $1\!:\!64$ ALU rate and \code{--use\_fast\_math} does not
apply to doubles. Second, the CFL wavespeed bound is time-invariant
and additively separable for every Hamiltonian here, so the
per-stage full-grid reduction can be deleted, not just fused. Third,
the SSP-RK3 final stage may write $V$ in place because the stencil
reads come from the second-stage buffer --- a dataflow fact about the
integrator, not about any one kernel. Fourth, at the grid sizes that
fit a control budget the whole field is L2-resident, so ghost-cell
materialization (the compiler's \code{concatenate}) and shared-memory
tiling are both strictly worse than letting the cache serve the
stencil overlap. Together these put our solver ahead on every case in
both precisions ($1.3$--$1.4\times$ at FP64, $1.7$--$16\times$ at
FP32) while the BRT is unchanged to the verification tolerance.

For the real-time deployment regime this paper targets, the
hand-written CUDA route wins on both axes --- it is faster on every
tested case \emph{and} it ships as a dependency-free C++/CUDA library
that links directly into a robot stack with no Python runtime, no JIT
warm-up on the first control tick, and deterministic per-tick
latency. The \code{torch.compile} route remains attractive on
different grounds: it is far easier to author and modify, and it
brings automatic differentiation for learning-in-the-loop. The two
are complementary, with the crossover governed by whether the
discretization exposes structure (a time-invariant CFL bound, an
algebraic weight identity, an in-place integrator stage) that a
hand-written kernel can exploit and a compiler cannot.

\subsection{Future work and applications}
\label{sec:discussion:future}

Online BRT computation at robot rates opens an application surface
that the historical offline-only regime closed off. We see four
near-term application classes that the rates in
Sections~\ref{sec:results}--\ref{sec:applications} make possible,
plus several solver-side extensions.

\emph{Safety filter for VLA and learned policies}~--- a BRT recomputed
each control tick is a natural deterministic guardrail downstream of
a vision--language--action stack or any learned controller: at every
tick the value function $V$ certifies the proposed action or the
gradient $\nabla V$ supplies a provably safe override, regardless of
how the upstream policy was trained. This is the
safety-filter pattern of~\cite{fisac2019safety,hsu2024safetyfilter}
without the requirement that the disturbance model be known at
training time.

\emph{ADAS and AV pre-crash control}~--- the 2D drifter and 3D Air3D
cases are the canonical primitives for ground-vehicle reach-avoid
against a worst-case adversary; an online BRT lets the safety
boundary track the actual current friction coefficient, road
geometry, and worst-case other-vehicle behavior rather than a
precomputed nominal.

\emph{Humanoid robotics}~--- contact-rich locomotion induces
fast-changing dynamics from terrain and payload; an online BRT can
certify balance and footstep regions against a worst-case ground
disturbance and worst-case push, recomputed under the current
estimated contact model.

\emph{Multi-agent dispatch under non-stationary
disturbance}~--- the wildfire-response demonstration in
Section~\ref{sec:applications} generalizes to any homogeneous
multi-agent dispatch problem under a time-varying flow field, e.g.,
search-and-rescue UAV fleets in shifting winds, surface-vehicle
fleets in tidal currents, or perimeter defense against incoming
threats with time-varying intercept geometry.

Solver-side extensions: persisting device buffers across the
per-tick solver calls in the closed-loop fleet demonstration (each
tick currently re-allocates and re-uploads the static flow field);
and Tensor Core kernels for FP16/BF16 stencil compute, combined with
mixed-precision error bounds, which could extend the real-time
envelope (Fig.~\ref{fig:envelope}) toward the $100\,$Hz regime on 4D
problems. The most direct test of the central proposition --- that
the boundary between offline and online BRT computation has moved
--- is a quantitative closed-loop study of an online-recomputed BRT
in a safety-filter role, measured against the same robot under
offline-precomputed and neural-approximator baselines in a regime
where the disturbance model shifts during the mission.

%% file: sections/conclusion.tex
\section{Conclusion}
\label{sec:conclusion}

We have presented a CUDA implementation of the canonical
WENO5+TVD-RK3+Lax--Friedrichs level-set Hamilton--Jacobi--Isaacs
solver that runs at robot control time scales on a single workstation GPU,
and on a Jetson AGX Thor embedded SoC at $31\,$W. The pointwise
output matches the Stanford ASL JAX reference \code{hj\_reachability}
on eight benchmarking cases at FP64 while running $2.0$--$4.6\times$
faster, and the solver outperforms a \code{torch.compile}
reimplementation of the identical scheme on every case in both
precisions --- the fused per-stage kernels exploit scheme-level
structure (a time-invariant CFL bound, an algebraically restructured
WENO5 normalization, an in-place final RK3 stage) that a
general-purpose tensor compiler cannot assume. At FP32 the solver
clears the $30\,$Hz bar with margin on the canonical 3D Air3D
($112\,$Hz) and 4D pursuit--evasion ($61\,$Hz) problems that have
historically been precomputed offline. An
in-kernel label-propagation extension delivers the $K$-agent
capture-basin Voronoi tessellation at the cost of a single combined
solve, which we demonstrated on a closed-loop multi-quadcopter
wildfire-response simulation under time-varying, non-uniform wind.
These rates put a deterministic, formally-guaranteed safety primitive
inside the inner loop of the autonomy stacks
--- ADAS/AV, humanoid robotics, and VLA-driven policies --- that
otherwise have no such guarantee available at deployment time.

%% file: sections/acknowledgment.tex
\section*{Acknowledgment}

The authors thank Claire Tomlin, Sylvia Herbert, Somil Bansal,
Jason J. Choi, and Donggun Lee for helpful discussions and pointers to
related work, and Donna Weekly, CEO of Mirror AI, for supporting this
work.

Generative AI was used extensively in preparing this article. Claude
(Anthropic), a large language model, assisted in drafting and revising the
text of all sections, in writing and optimizing the CUDA/C++ solver,
benchmarking, and figure-generation code, and in checking bibliographic
metadata and citation contexts. The authors directed the work, reviewed and
edited all AI-assisted text and code, re-ran every reported measurement, and
verified each reference against its source; they take full responsibility
for the paper's content.